# Data Classification With Multiprocessing

Anuja Dixit

*Department of Computer Science and Engineering*

*The Ohio State University*

Columbus, USA

dixit.89@osu.edu

Shreya Byreddy

*Department of Computer Science and Engineering*

*The Ohio State University*

Columbus, USA

byreddy.4@osu.edu

Guanqun Song

*Department of Computer Science and Engineering*

*The Ohio State University*

Columbus, USA

song.2107@osu.edu

Ting Zhu

*Department of Computer Science and Engineering*

*The Ohio State University*

Columbus, USA

zhu.3445@osu.edu

*Abstract* - Classification is one of the most important tasks in Machine Learning (ML) and with recent advancements in artificial intelligence (AI) it is important to find efficient ways to implement it. Generally, the choice of classification algorithm depends on the data it is dealing with, and accuracy of the algorithm depends on the hyperparameters it is tuned with. One way is to check the accuracy of the algorithms by executing it with different hyperparameters serially and then selecting the parameters that give the highest accuracy to predict the final output. This paper proposes another way where the algorithm is parallelly trained with different hyperparameters to reduce the execution time. In the end, results from all the trained variations of the algorithms are ensembled to exploit the parallelism and improve the accuracy of prediction. Python multiprocessing is used to test this hypothesis with different classification algorithms such as K-Nearest Neighbors (KNN), Support Vector Machines (SVM), random forest and decision tree and reviews factors affecting parallelism. Ensembled output considers the predictions from all processes and final class is the one predicted by maximum number of processes. Doing this increases the reliability of predictions. We conclude that ensembling improves accuracy and multiprocessing reduces execution time for selected algorithms.

I. INTRODUCTION

*A. Approach*

Modern AI consists of various tasks such as regression, prediction, etc. that collectively contribute towards making an intelligent system. One of the most important tasks is classification. There are numerous algorithms to approach this task with considerable accuracy. Each algorithm performs differently with different types of data. Performance of a classification algorithm is also affected by the hyperparameters used to train the model. Those hyperparameters also decide the number of resources used to complete the execution of the model. For any classification task, the best model is the one that consumes as few resources as possible but also produces the most accurate results with new or unseen data. It may take a series of executing various algorithms with combinations of hyperparameters until the best model is decided. Thus, the choice of hyperparameters is important for both deciding the resource consumption and efficiency of the model.

After the AI revolution in 2006, huge research is dedicated towards finding efficient ways to implement the classification algorithms, but one problem is consistent in all these years that is the time required to train the selected algorithm with appropriate hyperparameters on the specific data. Some might prefer to train the model using a subset of actual data to reduce the time required to decide the best combination of hyperparameters and later use that configuration to train the model with entire data. This approach might not give anticipated results because of various reasons like over-tuning the hyperparameters or selection of hyperparameters based on a biased subset. One way to minimize the time required to find the best combination of hyperparameters is to incorporate parallel processing while training and testing the model. This paper proposes using Python multiprocessing library to classify the data parallelly.

Python language is gaining popularity because of the flexibility and computational power it offers [review paper]. It also offers easy integration with other languages, making execution of external code module cost efficient. With a wide range of scientific libraries, Python also has dedicated libraries that help implement parallelism, multithreading, and multiprocessing. Whether to use multithreading or multiprocessing depends on the nature of the task we are trying to split. Multithreading is efficient when we are dealing with I/O bound tasks whereas multiprocessing gives efficient results while dealing with CPU bound tasks.

The approach presented in this paper mainly focuses on multiprocessing since the task of building and using the classification model is CPU bound. It is assumed that the model resides on the same machine as data. So, there is no I/O communication of any sort. Using multithreading in this case will give adverse effects due to the concept of Global Interpreter Lock (GIL) in Python. This is mutex lock used by threads in Python that allows execution of single thread at any given time. Only one thread can access the interpreter making it difficult to implement parallelism for CPU bound tasks. In the case of multiprocessing, each process has its own copy of interpreter and hence it can complete the execution parallelly with other processes giving better results specifically for CPU bound tasks.

Basic idea is to consider a classification algorithm and create a separate process for different configuration of hyperparameters for that process. Later, ensemble the output to generate the final prediction of classes. Since it is multiprocessing, processes can execute parallelly in less time as compared to serial execution of all those processes.

Ensembled output can help incorporate the benefits of all hyperparameters rather than focusing on a single configuration resulting in increased accuracy.

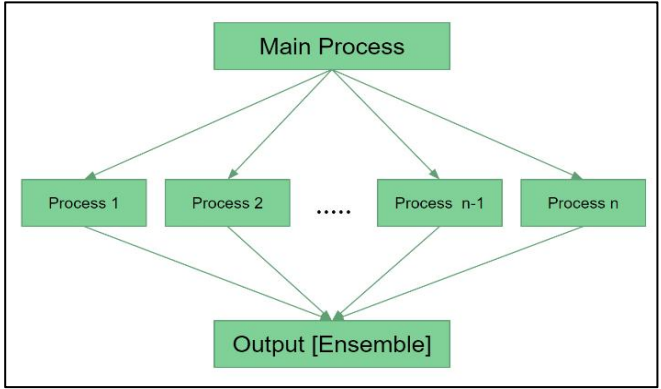

Fig. 1. Proposed approach

*B. Classification algorithms*

The proposed approach is implemented and tested with four classification algorithms:

1. K nearest neighbors (KNN)

   This algorithm finds the number of nearest neighbors (k) given in the model. We could use different distance metrics like Manhattan, Euclidian etc.… to find the k nearest neighbors. It picks out the label that is most repeated among these k neighbors and assigns that value to the unknown record.

2. Support Vector Machine (SVM)

   SVM tried to classify the entire dataset by fitting a specified kernel model to the data. The main goal of this algorithm is to fit the kernel to the model by maximizing the margin between the model and the data. This model allows some training errors to happen without exactly fitting this data, which results in overfitting the dataset.

3. Decision Tree

   This model makes use of two different and independent hyperparameters: minimum number of leaf nodes and depth of the tree. The algorithm keeps splitting the tree until it reaches the desired number of leaf nodes or the depth.

4. Random Forest

   This uses the number of decision trees specified in the algorithm. All these decision trees are independently run and the output of each of these trees are then ensembled to produce the final output for test data.

Clear explanation of different parameters we choose to train our dataset under different classification algorithms are explained the design section (III)

*C. Parallel processing:*

Cores in modern machines are independent and can execute multiple instruction at the same time. In multiprocessing, the main program is divided into smaller programs and those small pieces can be executed simultaneously to exploit parallelism with better CPU utilization. For any classification algorithm, every configuration of hyperparameters is independent of each other. Thus, it can be executed in parallel without being dependent on other processes. We propose the same approach to train multiple models at the same time and incorporate advantages of different configurations in the final classification model.

Parallel processing has some consequences as well. Processes need more resources to create and maintain as compared to threads. Thus, while incorporating parallelism in any code, considering this overhead is important.

*D. Python*

There are various classification algorithms available in Python sklearn library that are easily configurable and can be parallelized with the help of Python multiprocessing library. Each process can be assigned a configuration and the predicted results can be stored in a shared data structure. In this approach the data structure is a queue managed by manager class of multiprocessing. There is another way of using queue with multiprocessing but choosing the one managed by manager has some advantages which are discussed later in this paper.

Fig. 2. Data distribution w.r.t class attribute

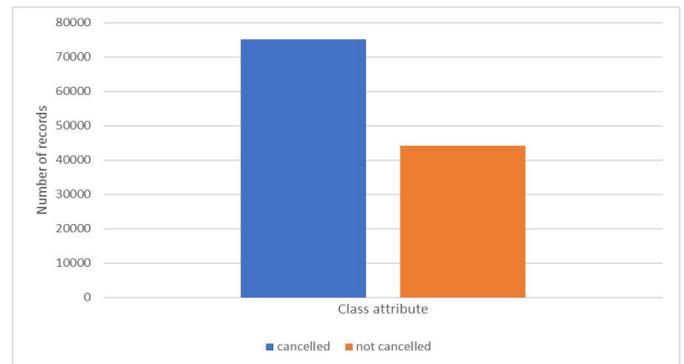

*E. Dataset used*

To implement our code and test our hypothesis we used a dataset available online from Kaggle [4]. This dataset contains records of users who have made reservations for a hotel. The attributes of this data include information that a user has entered while making a reservation. These attributes include information like number of adults, children, date of reservation etc.... There is a total of 32 of these attributes and 119,388 bookings. These attributes do not have any personal information of users like their credit card information, account number of their SSN. Among the 32 attributes one of them is the class attribute "is_cancelled" which is a binary value. The goal of any classifier is to make use of the given attributes and effectively predict this class attribute for a new incoming reservation. Distribution of number of records with respect to class attribute has been shown below in Fig 2

The paper is organized as follows. Next section discusses some prior work with multiprocessing in Python, section III has detailed design of the approach, section IV and V focus on results, evaluation and issues faced during working with the given approach. Section VI concludes the paper and includes some future work.

## II. RELATED WORK: LITERATURE REVIEW

We referred to a few papers that gave an idea of performance improvement by using multiprocessing in python. Reading a few articles online also gave us an insight into parallelization using Python multiprocessing. A review paper [1] gives a general idea about related work and various approaches to implement multiprocessing in Python. It also referred to a few useful resources that have implemented concepts which are an integral part of the proposed approach. Even though multiprocessing could be effective with executing code in parallel, the libraries that implement parallel programing could be vulnerable. There is a need to avoid data leakage and malicious attacks when it comes to using these publicly available libraries for implementation.

In the paper titled YouTube data collection using parallel processing [2], authors mention that they were able to achieve a 400% decrease in the processing time when parallel processing was incorporated into their work. Their aim was to collect the data from YouTube data APIs and store the data for analysis. To take advantage of multiprocessing, they created a function to submit and process the API requests. Then multiple processes were created, and each process completed API requests. The problem they were trying to solve was I/O bound. That gave an upper hand for results because the overhead caused due to I/O operation got distributed over multiple processes. Also, when a process was busy waiting for API response, the CPU can switch to other process increasing the throughput and performance. This gives some insights about the working of multiprocessing.

G. Heine, T. Woltron, and W. Alexander [3] deal with asynchronous streaming for monitoring public opinion using multiprocessing. It mainly focuses on asynchronous concurrent database writing methods with Python multiprocessing. This paper gave an insight into using queue as a communication medium between parent and child processes. Use of queue and parallelism greatly improved the performance while inserting data into Cassandra.

We found another paper [18] which compares the Aho-Corasick algorithm execution in parallel and serial. The authors concluded that the parallel execution of the algorithm gave better results in C as compared to the results in Python. The paper also mentions that Python may not be suitable to parallelize algorithms which are CPU bound as opposed to algorithms that are I/O bound. This paper is useful for expanding the scope of this project using other languages like C.

Multithreading in Python has a few conditions. One constraint is the presence of GIL due to which threads execute serially and we need to use multiprocessing if we are to implement parallel execution. In the paper Thread and Process Efficiency in Python [23], the authors have discussed the effects of GIL in detail. When we use multiprocessing instead of multithreading, some overhead occurs because processes need more start-up time as compared to lightweight thread. Thus, if we want to implement parallelism with multiprocessing, we need it to work with certain amount of data so that parallel execution using process is faster than the parallel execution using threads [23]. If we use large dataset, then the extra start-up time taken by the processes will be compensated and multiprocessing can execute faster than multithreading.

Techniques have been implemented to automate code generation and documentation with the help of python corpus[24]. Data scarping from the web has been done by combining a range of parallel corpus of python libraries. The baseline of this code generation and code documentation has been compared with machine translation produced by neural networks. Another reference [26] about Convolutional Neural Networks (CNN) compares parallel performance of multicore CPUs. It presents an idea of achieving speedup for training a model with multicore CPUs. This uses an image dataset and can give an insight into possible techniques for using our proposed approach for image classification.

There are numerous other resources related to the implementation of parallel processing in Python but due to time constraints, we decided to stick with the few that had concepts which are used in the proposed approach. We decided to narrow down the research to concepts specific to the multiprocessing class in Python, queue created with manager class, four classification algorithms that we are using, and related hyperparameters. We referred to some blogposts and articles [10][11][14] that shed light on issues like what is mentioned in the Challenges section (V).

Earlier research only focused on using multithreading for input output operations but there wasn't any analysis on how to import the performance of classification algorithm itself. They focused on reducing either time or improving accuracy, but this report uses techniques to incorporate both time reduction and accuracy improvement simultaneously. Thus, we wanted to try out our model which uses parallelism and ensembling to improve time and accuracy at the same time.

## III. DESIGN

As briefly mentioned in the introduction, the idea consists of configuring hyperparameters for each process and executing multiple such configurations in parallel to improve performance and efficiency of the classification model. Main motivation behind this approach is to minimize the computational time to classify the data and to find the best algorithm with the best parameters that classifies the unknown data with minimum error.

The configuration of a classification model varies as per the data it is dealing with. Thus, another motivation behind the proposed approach is to find an algorithm that gives the best results irrespective of the data it is dealing with. Since the ensembled output considers classification from all the configurations and chooses the class that is given by maximum number of processes, it combines the advantages from various configurations. Thus, we want to check if this approach helps us find such data independent algorithm or not.

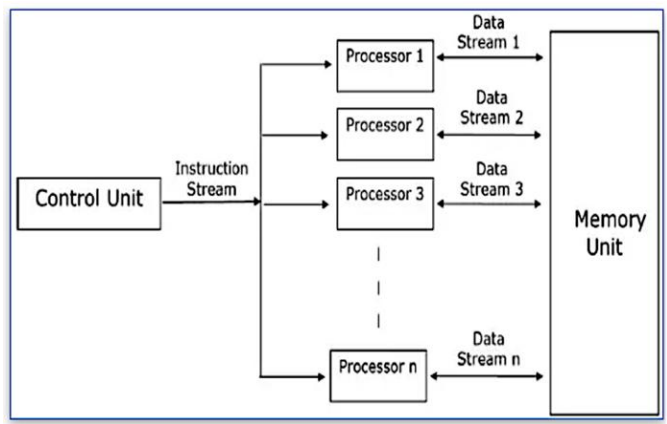

Fig. 3. Structure of parallel processing [1]

For implementing the classification algorithms, we are considering classifiers from Python sklearn library. To keep the results for serial and parallel execution comparable, we are considering the same function with same configurations to execute in serial and parallel. The only difference between the two executions is that for parallel processing, the model is split into small computational modules and those modules are executed in parallel using multiprocessing. The parallel model is doing the same amount of work as the serial model but since it is incorporating multiprocessing, we propose that the mentioned approach will complete the execution in less time as compared to the serial model.

*A. KNN classifier*

For implementing KNN classifier, we used KNeighborsClassifier from sklearn.neighbors in Python [9]. The hyperparameter considered is k or the number of nearest neighbors considered while classifying new data point. It is a general observation that accuracy of the model with same k changes with data. For a dataset where the classes are not well separated, smaller value of k gives more accurate results and larger k decreases the accuracy because of increased misclassification. Whereas, for the dataset where the classes are well separated, larger value of k avoids misclassification and increases accuracy. Thus, in both cases, using models with different k and deciding the final class by combining the output from multiple models will increase accuracy.

While we can achieve improved accuracy by engaging multiple models, it affects the execution time directly. The greater the value of k, the more time it takes to classify the data since the model needs to compare more distances to predict the output. Executing multiple models in parallel saves some time and utilizes the available resources to the maximum capacity.

There are multiple ways to approach the implementation of parallel execution model for KNN classifier. One way is to create a separate process for each value of k and run all the processes simultaneously. A major drawback of this approach is, with changing value of k, the number of resources required to complete the execution changes. We examined that by monitoring the resource utilization for different values of k. Even though the relation between the value of k and resource consumption is not linear or direct, generally, a model with lower value of k requires less resources as compared to a model with higher value of k. This creates an overhead on the main process while executing in parallel when separate processes are created for smaller k. This happens because the model with smaller value of k might not need the resource allocated to an individual process and creating and destroying the process consumes more resources than the actual model with small k.

To tackle this problem, we distributed the workload of individual processes by assigning multiple k to one process. By doing this we made sure that each created process has enough work to do that can consume the allocated resources and the overhead to create and destroy the process is negligible as compared to the execution time it is saving by working in parallel. We observed significant changes in the outcome in terms of execution time. Current working approach is as follows:

1. We focused on considering the results for k=1 through 20

2. Each created process will train the model using 4 values of k the distribution is done so that each process will have 1-2 smaller k that required comparatively less resources

3. Created 5 processes each with i = 1 through 5. Each process will train the model with value of k as i, i+5, i+10 and i+15

4. For serial execution, k = 1 to 20 are executed serially

*B. SVM classifier*

Sklearn library has a direct implementation of svm classifier [7]. The selected hyperparameter was the type of kernel used to determine the support vectors. Possible values for kernel are sigmoid, poly and linear. In this case, 3 processes were created, each implemented with a different kernel. For serial execution, the models with different kernel were executed one after the other.

*C. Decision tree classifier*

This classifier is implemented in sklearn as tree.DecisionTreeClassifier. It considers several parameters as an input [6] but, we chose two hyperparameters. One is min_sample_leaf which is the minimum number of samples required to be at leaf. Another parameter is max_depth which represents the maximum depth of the tree. These parameters can help avoid overfitting and increase the reliability of the prediction given by the decision tree.

The approach for parallel execution of decision tree is very simple. We created a list of values we want to train the model on, and a process is created for each value in the list. We followed the same approach for both parameters. In serial execution, classifier for each value is executed one after the other.

Main motivation of using decision tree classifier was its lightweight nature. Because the classifier is lightweight, implementing multiple configurations of this model won't be costly in terms of resources and each process will have enough work to perform which avoids the overhead problem discussed in KNN classifier. Using a single decision tree to predict the final output is a naïve approach and might not create very accurate results. But using more than one decision tree increases the reliability of the output along with the accuracy of classification. It also increases the robustness of the algorithm.

Using multiple decision trees and then considering the output given by maximum number of classifiers is very similar to using a random forest classifier. But this approach avoids a common problem of overfitting faced by the random forest classifier. Since we are using an extremely limited number of decision trees, there are very less chances of overfitting but we are still getting the advantage of better performance in terms of accuracy.

### D. Random forest classifier

This classifier is implemented in sklearn.ensemble Python module and accessed using RandomForestClassifier name [8]. Basic concept of random forest lies in decision tree. One random forest implements a given number of trees and ensembles the output to predict the final class of the data point.

There are two possible approaches for implementing parallelism with random forest. One is using the n_jobs parameter while creating the classifier function. That parameter decides the number of processes used for parallel execution of the code. In this approach, Python takes care of the multiprocessing and parallel execution. Another approach is to define the number of trees for each process and execute those processes parallelly. This approach implements multiprocessing for random forest from scratch instead of using the in-built parallelism in Python.

We decided to go ahead with the second approach of implementing the parallelism from scratch for fair comparison with other classifiers. We decided on an optimal number of trees [19]. For parallel execution, different processes executed the code with different number of trees whose total was equals to the decided number of trees. For parallel execution we have chosen a set of 64 and 66 trees where each set value is input into each process. For serial execution, only one process executed the random forest with decided number of trees. According to the number of trees assigned in parallel a total of 130 trees have been chosen for serial execution. Details of this approach are explained in the evaluation section (IV).

### E. Ensembling of output

Once the results for each of the classifier with their corresponding hyperparameters were obtained, they have been ensembled in such a way that the accuracy is improved when compared to the maximum output produced by the same classifier with a single set of hyperparameters. In ensembling we considered all the predictions produced by a classifier with different hyperparameters and chose the prediction that was produced maximum number of times. Thus, ensembling effectively predicts the correct label by combining the output for multiple weak classifiers.

## IV. EVALUATION

The evaluation of the proposed approach consists of comparing the time required to complete the execution for serial and parallel execution along with the accuracy of the model. A few other benchmarks include Recall, Precision and F1-score. Each of these benchmarks focuses on different things like false positives and true positives. Focus on recall, precision and F1-score depend on the nature of the problem we are dealing with, and these benchmarks are for reference purposes only. The focus of evaluation will mainly be on the execution time and accuracy of the model.

### A. Experiment setup

The dataset for hotel reservations was collected on 4/8/22. Our goal with this dataset is to accurately predict if a new incoming reservation is going to be cancelled or not depending on the existing dataset and its attributes. Making this prediction beforehand would help the company to know who would cancel and they could provide those people with incentives or more benefits so that they can change their plans. This would eventually give the company more profits.

The initial data collected was not clean and had missing values, outlier, duplicate records, and redundant attributes. Duplicate attributes were unnecessary and would in fact increase computation time and thus we removed all the duplicate records. We also removed the records with missing values for the class attribute which is is_cancelled as they would not be useful to us in any kind for prediction. Missing values of attributes were filled by the average of all the records for that attribute.

Attributes for building the classification model were chosen very carefully by finding the ones that were strongly correlated with the class attribute. Many tests like Chi-Squared [5] and ANOVA [20] were performed to choose the desired attributes.

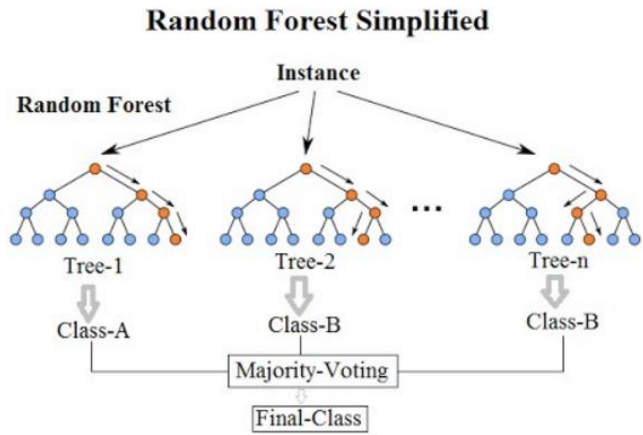

Fig. 4. Algorithm behind random forest

Only one among the attributes that had were strongly correlation was chosen as the other one would be redundant. All our chosen attributes did not vary a lot with their ranges, that was the reason why we did not normalize the attributes.

With the help of these attributes the data was split into training and test in the ratio of 70:30 with the help of sklearn libraries in python. Eventually these were passed into the classification algorithms with their respective parameters as specified in the design section (III)

The expectation for our model is that it outperforms serial execution time and accuracy. In real world situations we would like to reduce the generalization error for any of our predicted models. We would want to make sure that our model correctly predicts the class of a new incoming record. This can be done by training our model on a subset of the data called training data and predicting the labels on the remaining subset called test data. If our model correctly predicts most of the labels, then we can say that it would correctly predict when gone into production. This error value can be directly related to accuracy. The more the model accurately predicts test data labels, the less is its generalization error.

Below we have compared all the filtered classification algorithms and their performance with respect to serial and parallel with ensembling.

*B. KNN classifier*

Initially, we created a separate process for each value of k from 1 to 20. Even though the accuracy was increasing in the parallel execution, the time required for completing the parallel execution was more than the serial execution. After researching we figured out the problem that classifiers with the smaller values of k do not need many resources to complete the execution. So, creating a separate process for those configurations adds an overhead on the main process. Thus, we decided to move ahead with each process training more than one classifier. Each classifier will be trained on a different value of k.

We observed better results after we assigned multiple k values to one process. Computational time for parallel execution was significantly less than serial execution. Also, the accuracy of ensembled output was more than that of the maximum accuracy of serial output. These results show that the proposed approach was successful for KNN classification algorithm.

*C. SVM Classifier*

SVM classifier tries to fit the kernel specified in the hyperparameter to fit the model. It was observed that both time and accuracy were slightly lower for parallel when compared to serial. This is because the overhead was caused by SVM classifier. More details about this limitation have been explained the challenges section (V)

After running the algorithm with parallel and serial it was observed that parallel takes less time than serial but there was not much improvement in accuracy. Even though multiprocessing works for SVM, ensembling is not very helpful as the best results produced from serial model itself were good.

SVM uses the technique of maximizing its margin by including some amount of training error. This makes sure that the model does not overfit the data. Since SVM using the concept of maximum margin it is robust towards noise points in that data.

*D. Decision tree classifier*

We trained our dataset on decision tree classifier with two different hyperparameters: min number of leaf nodes and max depth. The set of minimum number of leaf nodes we used was [10,15,20,30,35,40]. Each time the decision tree was run with each value of leaf nodes.

The choice of number of leaf nodes is important because we do not want to choose a large number which results in splitting the records under each attribute. This could give us less training error but would eventually result in overfitting the data. We also do not want to choose the value less as that would not help us with generalization and can result in underfitting.

Max_depth parameter avoids overfitting. It terminates the algorithm if the depth of the tree is going beyond the maximum depth specified. The set of values we tried for the max_depth parameter was [5,7,9,11,13,15]. The results of the execution were as expected. Parallel computation took less time to as compared to the serial execution and the accuracy of the ensembled output was also higher for parallel execution. When we compared the results of using max_depth and min_sample_leaf parameters, we observed that using multiprocessing with min_sample_leaf achieves better results than using max_depth. Thus, we have included the results only for min_sample_leaf.

*E. Random forest classifier*

Since random forest by itself is a kind of ensembling algorithm with different decision trees as shown in Fig 4, applying our technique of ensembling by assigning the output of maximum number of classifiers would be useless. That was the reason we mainly focused on reducing the execution time for random forests by splitting the total number of trees in a random forest among multiple processes.

The number of trees in the random forest plays an important role while deciding the accuracy of the classifier. As per [19], after a threshold value of maximum number of trees is reached, we can see no improvement in the results. On the other hand, the algorithm consumes more resources for completing the execution. Until the threshold is reached, increasing the number of trees has a positive effect on the output of the classifier. As per the conclusion of [19], the optimal number of trees in random forest is 64-128. Considering all the factors, we decided to use 130 trees for random forest classifier.

As per our approach for parallel execution, we split 130 into several processes. Initially we decided to use 5 processes and split 130 accordingly. Almost all the processes dealt with less than 60 trees and the combined accuracy was less than the accuracy of serial execution. We thought of increasing the number of trees each process was dealing with. Since we had the constraint of 130 trees in total, we reduced the number of processes to 4. We were able to see improvement in the accuracy as compared to parallel execution with 5 processes. For exploiting the advantages of parallel computing, we decided the number of processes same as the number of cores i.e., 2. One process will create a random forest classifier with 64 trees and the other with 66 trees. That way we were able to respect the constraint of 130 trees in total. We were also able to meet the number of minimum optimal trees (64) as mentioned in [19].

For serial execution, only one process was executing a random forest with 130 trees. We observed some positive changes in performance after modifying our approach. The execution time was reduced substantially and the accuracy for ensembled output increased. For visualized results, please refer to Fig. 5., Fig. 6., and Fig. 7.

TABLE I. PROS AND CONS OF TRAINED ALGORITHMS

| Classification algorithm | Pros | Cons |
| --- | --- | --- |
| KNN | Simple and intuitive<br><br>No training step, so easy to predict. | Train through multiple K values<br><br>Slow and sensitive |
| SVM | Works best for models that are linearly separable | Doesn't perform well due to large dataset.<br><br>Time consuming |
| Decision Tree | Less effort for data preprocessing<br><br>Quick and efficient Performance is best for prediction. | Inappropriate for continuous values<br><br>Must rebuild with small changes to data. |
| Random Forest | Can easily be parallelized (per tree)<br><br>Reduced problem of overfitting due to many trees<br><br>Less variance than other methods | Slow<br><br>Many not perform as well as neural nets<br><br>No explicit formula for calculations |

### F. Results

Fig. 5. Time comparison between parallel and serial for KNN classifier

Time has been calculated using the built in time() function for python [21]. In the case of parallel execution, we started time calculation before creation of processes and ended it after each process finishes execution and joins. In the case of serial execution, we started the time before the initial classification algorithm starts and ended it after the last algorithm finishes execution. We did not include ensembling into time calculation as this would it a separate concept for increasing accuracy our model. As expected, parallel execution with multiprocessing is taking less time than serial.

When two processes are created to run the model in parallel, the time required to complete parallel execution is not reduced by 50%. This happens because multiprocessing takes time to create, start and join the process. Thus, even if the actual execution happens in less time, the main process needs to wait until all processes complete the execution and produce the results. This overhead is not present in the serial execution.

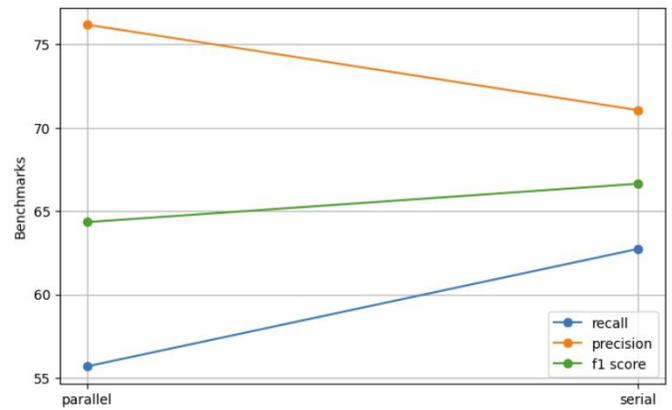

Fig. 6. Benchmark (recall, precision and f1 score) comparison for KNN classifier

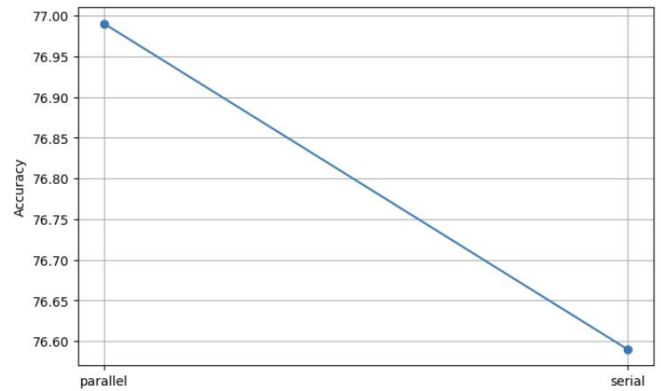

Fig. 7. Accuracy comparison for KNN classifier

Since this dataset contains imbalanced class attributes where there are not equal portions of both cancelled and not cancelled records, accuracy alone would not be a best measure for deciding on the best algorithm. Recall reduces the number of false positives. In this dataset the company would get more benefit by predicting that a booking would be cancelled than predicting that it would not be cancelled. Thus, from Fig. 6. it is seen that our parallel model with ensembling gives higher accuracy than the serial model.

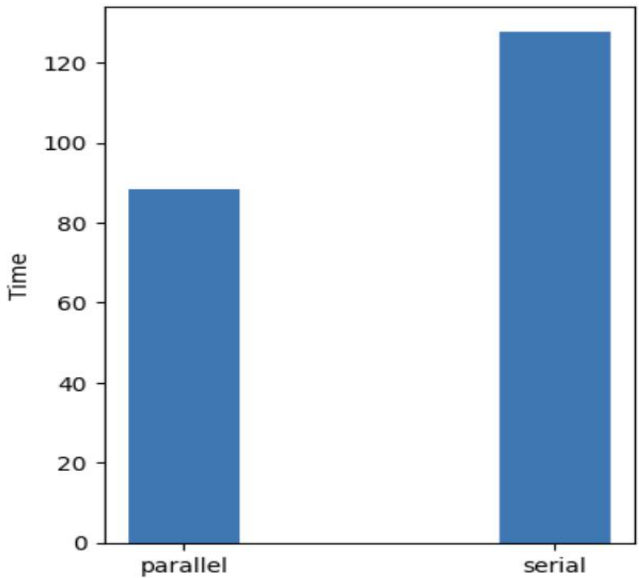

An ideal model needs to have both precision and recall close to 1. High scores of both recall and precision mean that the classifier is returning correct results for a lor its data. High recall and low precision mean that the labels predicted are incorrect. Low recall and high precision mean that even though the predicted labels are few, they are predicted correctly. Thus, low recall and high precision is better than high recall and low precision. Our parallel ensembled model achieves low recall and high precision when compared with serial model as shown in Fig 6.

Since parallel model uses ensembling and it assigns class that was given as output by most of the algorithms it is expected to produce more accurate prediction than serial model with only 1 classifier that produces max number of accurate results shown in Fig 7.

TABLE II. RELATIVE COMPARISION BETWEEN DIFFERENT CLASSIFIERS

| Classification algorithm | Benchmarks | | |
|---|---|---|---|
| | *Name* | *Parallel* | *Serial* |
| KNN | Time | 88.49 | 127.73 |
| | Accuracy | 76.99 | 76.59 |
| SVM | Time | 188.97 | 193.16 |
| | Accuracy | 72.21 | 74.03 |
| Decision Tree (min_sample_leaf) | Time | 3.2 | 8.2 |
| | Accurary | 79.75 | 79.45 |
| Random Forest | Time | 12.9 | 15.9 |
| | Accuracy | 82.06 | 82.04 |

Time is in seconds; accuracy is in percent.

The values of time and accuracy for each of the algorithms stated above varied a little bit but they were always better when compared to the serial execution.

## V. CHALLANGES

There were a few challenges while executing the model with the proposed approach. Those are listed as follows:

### A. Multithreading and multiprocessing

Threads are lightweight as compared to processes. Thus, we planned on using threads to implement parallel programming and improve performance of the classification model. But parallelism using multithreading works well only with I/O bound tasks. Execution of a classification algorithm is a CPU bound task making it challenging for multithreading to improve the performance. The results for parallelism using multithreading were worse than serial execution of the classification model particularly in terms of execution time.

The reason for the failure of parallelism using multithreading for a CPU bound task is the Global Interpreter Lock (GIL) in Python. It is a mutex lock used by thread that allows only one thread to access the interpreter at any instant. This directly counters the idea of parallelism and gives adverse results for execution time. Because of GIL, threads cannot execute in parallel and creating and destroying the threads creates an unnecessary overhead on the main process resulting in worse computational time.

In multiprocessing, a separate copy of interpreter is created for each process. This makes the execution of processes independent of each other resulting in parallel execution and improved results as compared to the serial execution.

### B. Execution on local system

We were working on a windows system using Jupyter notebook for executing the Python code and encountered an issue with the execution. Although the execution was proceeding, it wasn't printing any error message or output after a considerable amount of time. Upon research, we found that it has something to do with child process importing the appropriate module for execution. We tried to resolve the issue because we wanted to exploit the parallelism using 8-10 cores present on the local system, but because of the time constraints for the project and amount of remaining work, we decided to move forward with executing the code on Google Colaboratory (Colab). This impacted the results because the number of available cores on Colab was 2 whereas the local system could have executed the same code with 8-10 cores. Having multiple cores is advantageous for applications that incorporate parallelism since it allows execution of more instructions parallelly. Since we are using only 2 cores, the results do not show the expected improvement in performance. But looking at the constraints of resources, we can see considerable improvement in results when the model is executed in parallel.

### C. Variable results

As mentioned in the last challenge, due to system issues, we had you rely upon Colab for execution. The challenge was to reach a conclusion because every execution of the classifier produced different results due to caching. Thus, we needed to incorporate two actions while executing the code. The first one was to refresh the browser before every execution to avoid the problem of caching. Second was to execute the classifier multiple times before concluding anything. Second action was the main overhead which made execution challenging.

### D. Size of test dataset

As mentioned in [23], multiprocessing works better with larger datasets. We started working with the current dataset with around 120K records and after observing the multiprocessing behavior with the dataset, we found another dataset [25](dataset B). Dataset B has a structure different from that of the dataset we are using for the current project. It is an image dataset with 12,000 images. Each image corresponds to one record in the dataset and each record has 16,384 attributes and 1 class attribute. This makes it an interesting dataset to work with. But because we were working on Colab, we needed to upload the data every time we reconnected with the run time environment. Uploading this dataset B took a few hours and doing this every time after reconnecting was not feasible because of the tight schedule we were working with.

### E. Shared data structure

In the proposed approach, each process is executing the model and performing classification separately. In the end, to generate the final ensembled output, the main process needs the result form all child processes. To achieve this, inter-process communication is required so that child processes can convey the results to the main process. There were multiple approaches to implementing this. One of the approaches was to use a shared data structure that is accessible by the main process as well as all the child processes.

We finalized using queue data structure that is available in the multiprocessing Python module. That way we were sure that this data structure is compatible with multiprocessing and inter-process communication. The disadvantage of using the queue from the multiprocessing module is that it takes time to copy the results to the queue and increases the execution time while using multiprocessing. This happens because before the results are copied to the queue, they are pickled (converted to byte stream) and the pickled results are then flushed on to the underlying pipeline. Thus, the main delay happens because of the pickling and unpickling process of the object. Also, queues in multiprocessing use a synchronization primitive or lock so that only one process can access the queue at one time.

The solution to this problem is to use a queue created with a manager present in the multiprocessing module. A queue created using a manager is controlled by the manager and the manager makes sure that the queue is always accessible to all the processes. The output doesn't need pickling when we are using the manager queue. That reduced the time for parallel execution substantially.

### F. Execution time for SVM classifier

For multiprocessing with SVM classifier, we decided to use the type of kernel as a hyperparameter to tune and improve the performance. The possible values for the parameter are 'poly', 'sigmoid', and 'linear'. The challenge in this case was the execution time of the classifier with linear kernel. The classification algorithm was completing the execution for poly and sigmoid kernel in approximately 30 minutes when trained with the entire dataset. But, for linear kernel, it took more than 3 hours to complete the execution which was not efficient considering the time constraints we had for working on the project.

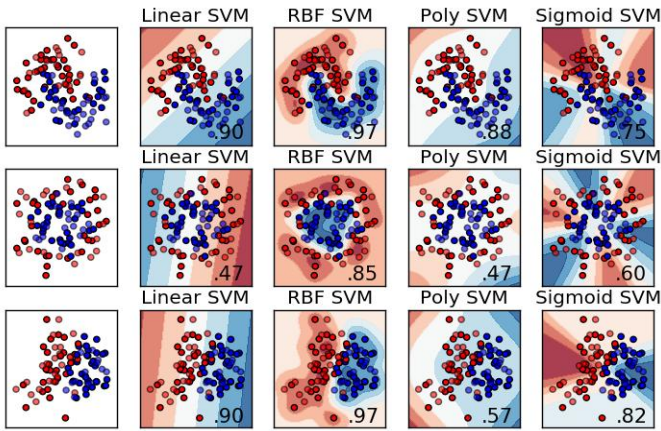

Fig. 8. Complexity of SVM for different kernels [22]

As seen in Fig. 8 SVM is very complex to fit the specified kernel to the given data. Thus, it is expected that it took a lot of computation time for such a complex algorithm and with nearly 120K records of dataset.

To solve this problem of exceptional execution time, we decided to sample the dataset. Selecting an accurate sample size is crucial since we do not want to miss significant information of the dataset. if we chose a sample size that is too small then we might lose some important details about the dependencies between features of the dataset and the class attribute. If the sample size is too big then there would not be any difference in the execution time and eventually no use of sampling.

The most easy and effective sampling technique that we used here is random sampling [27]. By using random sampling, we sampled a subset of 10,000 records to train the model instead of using the complete dataset. By considering a subset of dataset, we were able to get the results in reasonable time.

## VI. CONCLUSION

### A. Conclusion about the approach

Choosing attributes for classification plays a crucial role so that we do not overfit or underfit our dataset. A few hyperparameters show threshold based behavior meaning the performance of the algorithm doesn't improve by changing a value of the parameter after it hits some value i.e., threshold. This happened where there is a possibility of the parameter to take multiple values like k in KNN and number of trees in random forest. The performance of KNN gets worse after some value of k. Thus, it is important to choose the right value for hyperparameters.

After comparing results of all the classification algorithms, it was observed that random forest gives higher accuracy when compared with other classification algorithms tested. But KNN gives maximum benefit for our parallel and ensembled mode. The execution time for KNN gets reduced by 39.24 seconds for multiprocessing and accuracy increases by 0.4% according to Table 1. We can see positive results for decision tree algorithm as well. Thus, we conclude that ensembling improves accuracy and multiprocessing reduces execution time for selected algorithms.

In the case of SVM classifier algorithm, the execution time was reduced, but the accuracy of the ensembled output was less than the maximum accuracy from serial execution. Thus, for SVM, we conclude that ensembling the output after parallel execution does not work well and individual kernel can give better results.

Since different computers have different configurations like the number of cores, CPUs, the execution of our proposed model with give different results on different computers but they should be consistent with the performance. Irrespective of the configuration parallel model with multiprocessing should take less time when compared to serial model. Additionally, ensembling the output should improve the accuracy when compared with the original output with maximum accuracy.

### B. What we learned

The proposed approach needed some research to implement the execution strategy. There were two subjects of focus for this project. First was selecting appropriate classification algorithm and find the configuration of proper hyperparameters that helps to consider all the advantages offered by parallel computing. The other was to study multiprocessing and understand how it works in Python. This section below gives an overview of what we learned while working on this project.

1. Classification algorithms and hyperparameters:

    The goal of the project is to implement data classification with multiprocessing. For that we needed to find algorithms that can be split into multiple processes and execute parallelly which would be possible if we knew how different algorithms work. As a part of the project, we researched the working of the algorithms to decide the parallel computing component of the classifier. We also learned about different hyperparameters corresponding to each algorithm while deciding the configuration for parallel execution.

2. Multiprocessing and multithreading in Python:

    During the lectures, we learned about multiprocessing and multithreading. So, we were aware of the concepts, but we needed a clear understanding of implementing those in Python if we were to implement the project. Initial plan was to use multithreading. Thus, we researched multithreading in Python, how to achieve it and what are the drawbacks. While doing that, we learned about GIL, and understood the concept of mutex taught in the lectures more clearly. To go ahead with the project, we researched about multiprocessing in Python and studied what is better, multiprocessing or multithreading, as per the nature of the task (I/O bound and CPU bound). We understood the reasons for multiprocessing in Python being more useful in the case of tasks that are CPU bound.

3. Factors affecting parallel processing:

    While we compared the results of different classifiers, we studied factors that affect parallel processing. Parallelism is affected by choice of classification algorithm, hyperparameters and the size of the dataset we are dealing with. Algorithms like SVM are not suited for parallelization based on hyperparameters. From KNN algorithm, we got to learn the effects of hyperparameters on parallelism. And generally, we learned that with smaller datasets, parallel processing doesn't work well. Everything summarized to resource utilization. If an algorithm, after parallelizing, consumes enough resources then implementing multiprocessing is effective. Otherwise, the implementation of parallel execution will be an overhead on the main process resulting in adverse effects.

### C. Future work

A paper [18] shows that parallel implementation of one algorithm using multiprocessing in C was more efficient than the implementation of the same algorithm using multiprocessing in Python. Thus, future work of this project includes studying data classification with multiprocessing in C.

There is scope in future to try different ensembling techniques to improve the accuracy of a model to a much higher ranger. Techniques like boosting which uses a sample of features without replacement and training records with replacement to fit the model. Features are selected without replacement to ensure that no single feature is dominating the output. Adaboost is another technique which iteratively updates the prediction of the training data by focusing on records that were previously misclassified. We could also ensemble output from different classifiers instead of a single classifier with multiple hyperparameters.

Considering the constraints we had for the project, we worked with only one dataset for checking the approach. Future work includes working with multiple large datasets to check the consistency of the results with each classifier.

For this project, we are considering multiple configurations of a single classifier to execute in parallel and then to ensemble the output. This could be modified to use multiple classifiers with a few configurations and executing them in parallel and then the output from all classifiers will be ensembled. Eventually the effects of this approach on the execution time and accuracy could be analyzed.

In our model we have not addressed the concerns for threat to vulnerability. Some work in future could relate to the kind of attacks that multiprocessing would be prone to be involved in. Multiprocessing could be analyzed from the perspective of security issues.

In addition, we can also extend the results to more diverse areas in the future. In the context of IoT and heterogeneous computing systems, multiprocessing can play a significant role in enhancing the performance of data processing tasks. Studies have shown that leveraging concurrent communication methods and heterogeneous data sources, as discussed in [28-40], can significantly improve the efficiency and scalability of data classification systems. This is particularly relevant when dealing with large-scale IoT networks where data from various devices needs to be processed simultaneously.

Furthermore, the integration of multiprocessing techniques with advanced machine learning models, particularly in the context of low-power communication [41-43] and edge computing [44-47], offers promising avenues for research. Efficient data processing and classification on edge devices can lead to more responsive and intelligent IoT systems.

Future explorations could delve into the realm of secured protocols for IoT devices in tactical networks [48-49], ensuring the safety and integrity of data during multiprocessing tasks. The integration of multiprocessing with physical-layer security measures, such as message authentication for ZigBee networks [50], could be investigated to enhance the overall security of data classification systems.

Additionally, exploring the implications of multiprocessing in smart health applications [51-53] and sustainable energy systems [54-55] could lead to innovative solutions for real-time data analysis and decision-making. This includes the utilization of multiprocessing for efficient energy management in electric vehicle networks [56] and enhancing air quality management systems [57] through real-time data processing.

With the advancement of machine learning and AI in next-generation wireless networks [58], the potential of multiprocessing extends beyond traditional data classification tasks. The exploration of high-granularity modulation techniques for OFDM backscatter [59-60] and the development of versatile MIMO backscatter systems [61] present opportunities for multiprocessing in advanced communication systems.

Future studies could also focus on enhancing the robustness of multiprocessing systems against sophisticated cyber attacks, such as obfuscation-robust character extraction [63] and deep learning-guided jamming attacks [64]. Additionally, the integration of multiprocessing with chatbot systems and cryptocurrency platforms [62,65] could lead to more efficient and scalable solutions in these domains.

In summary, the future of data classification with multiprocessing is not limited to enhancing performance in Python but extends to a wide array of applications in IoT, security, and smart systems. The intersection of multiprocessing with emerging technologies opens new pathways for innovative research and practical implementations.